\documentclass[lettersize,journal]{IEEEtran}
\usepackage{amsmath,amssymb,amsfonts}
\usepackage{algorithmic}
\usepackage[ruled]{algorithm2e}
\usepackage{array}
\usepackage{textcomp}
\usepackage{stfloats}
\usepackage{url}
\usepackage{color}
\usepackage{verbatim}
\usepackage{graphicx}
\usepackage{multirow}
\usepackage{array}
\usepackage{booktabs}
\usepackage{graphicx}
\usepackage{parskip}

\usepackage[numbers]{natbib}
\usepackage{bm}
\usepackage{mathtools}

\usepackage{subfigure}
\usepackage{makecell}
\usepackage{enumitem}

\usepackage{bibspacing}
\graphicspath{ {./fig/} }

\usepackage{flushend}

\usepackage{stackengine}

\def\BibTeX{{\rm B\kern-.05em{\sc i\kern-.025em b}\kern-.08em
    T\kern-.1667em\lower.7ex\hbox{E}\kern-.125emX}}
\usepackage{balance}
\begin{document}

\title{Semantic Chain-of-Trust: Autonomous\\ Trust Orchestration for Collaborator\\ Selection via Hypergraph-Aided Agentic AI} 
\author{
Botao~Zhu,~\IEEEmembership{Member,~IEEE}, Xianbin~Wang,~\IEEEmembership{Fellow,~IEEE}, and Dusit~Niyato,~\IEEEmembership{Fellow,~IEEE}
\thanks{
B. Zhu and X. Wang (Corresponding author) are with the Department of Electrical and Computer Engineering, Western University, Canada. D. Niyato is with the College of Computing and Data Science, Nanyang Technological University, Singapore.}
}

\maketitle

\begin{abstract}
The effective completion of tasks in collaborative systems hinges on task-specific trust evaluations of potential devices for distributed collaboration. Due to independent operation of devices involved, dynamic evolution of their mutual relationships, and complex situation-related impact on trust evaluation, effectively assessing devices' trust for collaborator selection is challenging. To overcome this challenge, we propose a semantic chain-of-trust model implemented with agentic AI and hypergraphs for supporting effective collaborator selection. We first introduce a concept of semantic trust, specifically designed to assess collaborators along multiple semantic dimensions for a more accurate representation of their trustworthiness. To facilitate intelligent evaluation, an agentic AI system is deployed on each device, empowering it to autonomously perform necessary operations, including device state detection, trust-related data collection, semantic extraction, task-specific resource evaluation, to derive a semantic trust representation for each collaborator. In addition, each device leverages a hypergraph to dynamically manage potential collaborators according to different levels of semantic trust, enabling fast one-hop collaborator selection. Furthermore, adjacent trusted devices autonomously form a chain through the hypergraph structure, supporting multi-hop collaborator selection. Experimental results demonstrate that the proposed semantic chain-of-trust achieves 100\% accuracy in trust evaluation based on historical collaborations, enabling intelligent, resource-efficient, and precise collaborator selection.

\end{abstract}
 


\section{Introduction}

\IEEEPARstart{W}{ith} the escalating complexity of both applications and the proliferation of interconnected systems, it has become increasingly infeasible for individual resource constrained devices to independently execute computing tasks due to their limited computational power and energy resources~\cite{9718357}. To overcome such limitations, recent research has focused on a new collaborative paradigm that leverages distributed peer resources to enable collaborative task execution~\cite{11048876}. This paradigm emphasizes coordination among devices through resource sharing to enhance the overall system’s task-processing capability and efficiency. Distributed collaborative computing has seen broad adoption across industrial sectors such as smart manufacturing, smart cities, and e-health, offering substantial benefits including reduced latency, improved resource efficiency, and enhanced service quality~\cite{10594714}. These advantages underscore its potential as a foundational enabler for next-generation collaborative systems.

Selecting reliable collaborators is essential for ensuring successful task execution in cooperative systems. Trust has recently been proposed as a holistic metric for characterizing a collaborator’s reliability, which is defined as the task owner’s expectation of a collaborator’s ability to complete a given task~\cite{chain_of_trsut}. Existing research has introduced various methods to evaluate the trustworthiness of collaborators. Some studies evaluate collaborator trustworthiness by examining behavioral indicators such as task completion rate, response latency, and error rate, while others leverage social and relational attributes, e.g., neighbor feedback and social connectivity, to assess collaborator trustworthiness~\cite{10223275}. However, these approaches typically produce a binary judgment--trustworthy or untrustworthy--which fails to accurately reflect the varying levels of device trustworthiness in dynamic collaborative systems. To address this limitation, we propose a concept of semantic trust in this paper to characterize device trustworthiness in a multi-dimensional and semantically meaningful manner. This new concept enables a more nuanced assessment that accurately captures the reliability of devices within dynamic, situation-dependent collaborative environments. Semantic trust evaluation aims to enable efficient selection of trustworthy collaborators under complex conditions. To improve selection efficiency, devices can be connected according to their semantic trust relationships so that trustworthy collaborators can be rapidly screened based on those associations when a task arrives. Building on this concept, we further propose a semantic chain-of-trust model, where devices are linked through semantic trust relationships to form a sequential chain that can dynamically adapt to changes in device trustworthiness. However, since devices operate independently and their trustworthiness evolves dynamically, several challenges must be resolved to realize the proposed semantic chain-of-trust.

\textit{How to efficiently perform semantic trust evaluation for collaborators?}  
To evaluate the semantic trust of collaborators, each device collects trust-related data from its collaborators, such as packet loss rate, task completion rate, execution latency, error rate, resource utilization, and task quality. The device then analyzes these raw data to extract semantically meaningful trust features and constructs a multi-dimensional semantic trust representation for each collaborator. This data collection and analysis process is complex and incurs substantial communication and computation overhead~\cite{10223275}, which is particularly costly for resource-constrained devices. Therefore, two key considerations must be addressed to achieve effective semantic trust evaluation. First, the semantic trust evaluation process must not interfere with normal task execution. Devices should be capable of autonomously identifying their idle periods and performing data collection, analysis, and semantic trust updates only during these periods, thereby maximizing resource utilization. Second, collaborators’ semantic trust must be dynamically updated to reflect changes in their behavior over time. Consequently, devices require on-device learning capabilities to autonomously learn the up-to-date semantic trust of collaborators from ongoing collaborations. Large AI model-driven agentic AI systems have recently attracted significant attention due to their emerging advantages. These include autonomous state awareness~\cite{10558819}, which enables agents to monitor and interpret environmental changes in real time, and continuous learning~\cite{10638533}, allowing agents to incrementally update their knowledge through ongoing interactions. Therefore, agentic AI holds significant promise for efficient semantic trust evaluation.

\textit{How to design an effective mechanism to dynamically chain distributed devices?} Because devices operate independently and their trust levels vary, devices must be capable of autonomously forming a chain that dynamically adjusts to trust change. To this end, the following two key issues must be addressed. First, a structured tool capable of representing multi-level semantic trust relationships between a device and its potential collaborators is needed. Each device typically interacts with multiple collaborators whose trust levels may vary significantly. Grouping collaborators with similar trust semantics into the same category enables devices to manage heterogeneous trust relationships more effectively in complex environments. Therefore, a tool supporting both semantic representation and hierarchical trust organization is essential. Second, a mechanism is needed to enable the autonomous formation of a dynamic chain among devices. This self-organizing chain must be formed without centralized control, allowing devices to autonomously establish associations based on their perceived semantic trust toward potential collaborators. As device trust continuously evolves, the chain must be able to expand, break, or reconfigure over time, necessitating a mechanism capable of supporting such autonomous and dynamic restructuring. Compared with conventional graphs, hypergraphs can organize nodes into multiple association sets through distinct hyperedges~\cite{10546264}. Each hyperedge can be annotated with labels that encode the semantic properties of its associated group. Additionally, hypergraphs support flexible topological evolution, allowing nodes and hyperedges to be added or removed dynamically. With their expressive representation of complex relationships and inherent adaptability, hypergraphs serve as an effective means for constructing a dynamic, self-organizing chain among devices.

Building on the above discussion, this study employs agentic AI and hypergraphs to realize the proposed semantic chain-of-trust. The core idea is to deploy an agentic AI system on each device, enabling autonomous execution of all operations required for evaluating the semantic trust of collaborators. Meanwhile, each device leverages a hypergraph to manage its collaborators according to different levels of semantic trust. Moreover, adjacent trusted devices autonomously form a chain through the hypergraph structure, thereby supporting cross-device collaborator selection. The main contributions of this paper are summarized as follows. 

\begin{itemize}[leftmargin=*]
    \item  We propose semantic trust for the first time to enable accurate multi-dimensional evaluation of collaborators, and propose the semantic chain-of-trust to facilitate rapid collaborator selection.

    \item We employ agentic AI to construct a framework for evaluating the semantic trust of collaborators, enabling fully autonomous and intelligent trust assessment through collaboration among agents powered by large AI models.

    \item We leverage the advantages of hypergraphs to establish a dynamic management framework for collaborators based on semantic trust and to form a distributed trust chain among devices, thereby efficiently supporting cross-device collaboration.
\end{itemize}

\vspace{-0.15 in}
\section{Related Work}
In recent years, with the rapid advancement of AI technologies, various AI-based trust evaluation approaches have been proposed for wireless systems. In \cite{10104094}, the authors proposed a trusted collaboration framework for mobile edge computing-enabled virtual reality video streaming, where trust is evaluated by combining direct historical interactions with indirect recommendations from other nodes. A multi-agent deep reinforcement learning approach is then employed to optimize collaborator selection. In \cite{11161582}, the authors presented a Siamese-enabled framework for continuous trust evaluation, in which the attributed control flow graph is used to capture the semantic information of collaborators’ communication and computing resource attributes. The authors in \cite{10877713} established a device trust model based on typical interactions in communication, security, and related behaviors, and integrated an artificial neural network to identify whether a device behaves maliciously or normally. However, these trust evaluation methods rely on machine learning techniques that require large amounts of training data, which limits their applicability in scenarios that demand genuine intelligence and adaptability. Large AI models have attracted significant attention in trust evaluation due to their powerful self-learning capabilities and minimal reliance on training data. In \cite{chain_of_trsut}, the authors presented a progressive chain-of-trust framework that leverages generative AI to perform stage-by-stage, context-aware trust evaluation, enabling efficient and reliable collaborator selection in dynamic environments. In \cite{10945110}, the authors introduce a hierarchical cyber-attack attribution framework and employ a large language model (LLM) to extract and populate the framework’s content from attribution reports. The result demonstrates that LLM can effectively cover the attribution of the primary stages of network intrusions.
While current methods provide partially intelligent trust evaluation, they do not encompass all required operations and therefore fall short of true intelligence. To overcome this, we introduce the semantic chain-of-trust model, realized with agentic AI and hypergraphs, achieving fully intelligent trust assessment for effective collaborator selection. Details are presented in the following section.

\section{Semantic Chain-of-Trust}

The proposed semantic chain-of-trust model is illustrated in Fig.~\ref{semantic-chain}. In this model, each device is equipped with an agentic AI system, which autonomously evaluates potential collaborators by collecting relevant data, assessing their semantic trust, dynamically managing them based on trust levels, and performing task-specific resource evaluation. Adjacent trusted devices autonomously form a chain-like structure, constructing a cross-device hypergraph that enables multi-hop collaboration. In the following subsections, we present the system model, the hypergraph and agentic AI framework, and the implementation of the semantic chain-of-trust.

\vspace{-0.1 in}
\subsection{System Model}
\label{trust_model}

In a collaborative system with a set of devices $B = \{b_1,\dots, b_I\}$, each device periodically collects the communication- and computation-related data generated by others during past task collaborations and leverages these data to infer their trust over a recent time window. In this work, historical collaboration-based trust is quantified along four dimensions: response latency, task transmission success rate, execution speed, and task computation success rate. This trust representation is extensible and can be augmented with additional metrics when required. When device $b_i$ acts as the task owner, it initiates a face recognition task $C$ with the parameters \{``size" : ``500 MB", ``collaborator willingness" : ``yes", ``minimum CPU requirement" : ``2 GHz"\}. The task $C$ contains a set of images and aims to count the number of individuals appearing in these images. Note that the task parameters can be extended as needed. The task owner $a_i$ selects a trustworthy collaborator to execute the task $C$ by evaluating each candidate’s semantic trust, including historical collaboration-based trust and resource-based trust.

\vspace{-0.1 in}

\subsection{Hypergraphs and Agentic AI}
\begin{figure*}[!]
\centering
\includegraphics[scale=0.95]{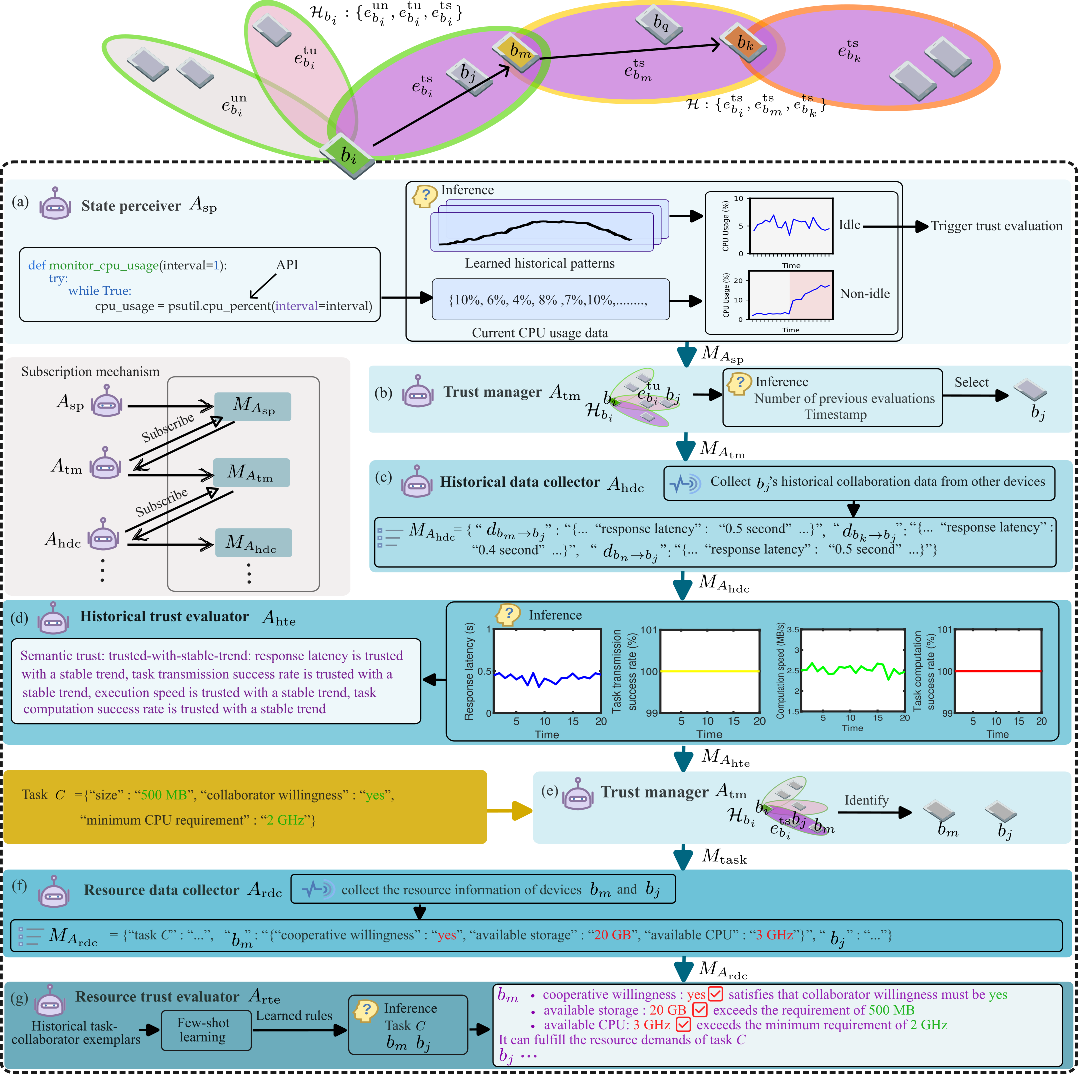}
\caption{The proposed semantic chain-of-trust model. Workflow of agents on device $b_i$ for trust evaluation involves:  
$(a)$ Agent $A_{\text{sp}}$ determines the timing of trust evaluations. $(b)$ Agent $A_{\text{tm}}$ decides which devices require trust evaluation. (c) Agent $A_{\text{hdc}}$ collects historical collaboration data of the selected device. (d) Agent $A_{\text{hte}}$ evaluates the historical collaboration-based trust of the selected device. (e) Agent $A_{\text{tm}}$ updates the local trust hypergraph $\mathcal{H}_{b_i}$; Agent $A_{\text{tm}}$ identifies the potential collaborators. (f) Agent $A_{\text{rdc}}$ collects the potential collaborators' resource information. (g) Agent $A_{\text{rte}}$ evaluates the potential collaborators’ resource trustworthiness.}
\label{semantic-chain}
\end{figure*}

\textit{1) Hypergraphs}: A hypergraph can comprise an arbitrary number of nodes and hyperedges, where each hyperedge can connect a subset of nodes to represent complex relationships. 
This research considers two types of hypergraphs: local hypergraph and global hypergraph. Each device maintains a local hypergraph to manage its potential collaborators. For example, in Fig.~\ref{semantic-chain}, device $b_i$'s local hypergraph is defined as
$\mathcal{H}_{b_i} = (B_{b_i}, E_{b_i}, W_{b_i})$. Here, $B_{b_i}$ is the set of all potential collaborators of device $b_i$. $E_{b_i} = \{e_{b_i}^{\text{un}}, e_{b_i}^{\text{tu}}, e_{b_i}^{\text{ts}}\}$ is the set of hyperedges, where each hyperedge represents a subset of potential collaborators. $W_{b_i} = \{w_{b_i}^{\text{un}}, w_{b_i}^{\text{tu}}, w_{b_i}^{\text{ts}}\}$ denotes the set of hyperedge weights, each encoding a particular trust semantics. Specifically, $w_{b_i}^{\text{un}}$, $w_{b_i}^{\text{tu}}$, and $w_{b_i}^{\text{ts}}$ correspond to untrusted, trusted-with-unstable-trend, and trusted-with-stable-trend, respectively. The trusted-with-stable-trend groups from all devices are then connected to form the global hypergraph $\mathcal{H} = (B, E, W)$, where $E = \{\dots,e^{\text{ts}}_{b_i}, e^{\text{ts}}_{b_m}, e^{\text{ts}}_{b_k}\dots\}$, and 
$W = \{\dots,w^{\text{ts}}_{b_i}, w^{\text{ts}}_{b_m}, w^{\text{ts}}_{b_k},\dots\}$.

\textit{2) Agentic AI}:
Each device $b_i$ is equipped with an agentic AI system implemented using MetaGPT~\cite{hong2024metagpt}, a general-purpose agent framework that supports multi-role, multi-task collaboration. The agentic AI system breaks down the complex trust evaluation into multiple smaller subtasks, each handled by an LAM-enabled specialized agent. Following the trust evaluation workflow, we create six agents with distinct roles, and initialize each agent with specific knowledge and skills tailored to its responsibilities. These agents work cooperatively to complete the semantic trust evaluation. A detailed description of the six agents is presented below.

State perceiver $A_{\text{sp}}$: This agent continuously monitors CPU activity to detect device idleness and, once idleness is confirmed, initiates the subsequent trust evaluation process.

Trust manager $A_{\text{tm}}$: It manages device $b_i$'s collaborators into hierarchical groups according to their semantic trust.

Historical data collector $A_{\text{hdc}}$: It invokes external modules to collect historical collaboration data of collaborators under evaluation from other devices.

Historical trust evaluator $A_{\text{hte}}$: It infers the semantic trust of collaborators under evaluation using the collected historical collaboration data.

Resource data collector $A_{\text{rdc}}$: When device $b_i$ generates a task $C$, $A_{\text{rdc}}$ invokes external modules to collect the current available resource data from all potential collaborators.

Resource trust evaluator $A_{\text{rte}}$: Based on the available resource data of potential collaborators and the requirements of the task $C$, it conducts a task-specific trust evaluation for each potential collaborator.

\subsection{Semantic Chain-of-Trust Implementation}

Trust evaluation comprises two components: assessing collaborators’ historical collaboration-based trust and evaluating their current resource suitability for the specific task. The task owner combines the outcomes of these components to identify a reliable collaborator for task execution. Within each component, specialized agents execute trust evaluation subtasks following a streamlined workflow. Agents interact through a subscription-based communication mechanism, where each downstream agent subscribes to the outputs of the upstream agent. Following the model context protocol (MCP), agents exchange messages in a structured format. Additionally, some agents interact with external modules via application programming interfaces (APIs). The following provides a detailed description of the workflow executed by the agents on device $b_i$ for performing historical collaboration-based trust evaluation on its collaborators.




\textbf{1) Historical Collaboration-Based Trust Evaluation}

\vspace{-0.03 in}
\textit{Step 1: Agent $A_{\text{sp}}$ determines the timing of trust evaluations}. To optimize the utilization of distributed resources, device $b_i$ conducts historical collaboration-based trust evaluation for collaborators only during idle periods. To this end, agent $A_{\text{sp}}$ operates in an ``observe–infer–act'' cycle. It continuously monitors device $b_i$’s CPU activity via API calls. However, relying on a single instance of low CPU utilization to determine idleness may yield inaccurate results. Instead, agent $A_{\text{sp}}$ infers the state of device $b_i$ by combining recent CPU utilization data with learned historical CPU activity patterns, as shown in Fig.~\ref{semantic-chain}~(a). Once device $b_i$ is confirmed idle, agent  $A_{\text{sp}}$ immediately outputs a message $M_{A_{\text{sp}}}$ to trigger the trust evaluation process.

\textit{Step 2: Agent $A_{\text{tm}}$ decides which devices require trust evaluation}. Since agent $A_{\text{tm}}$ subscribes to messages from agent $A_{\text{sp}}$, upon receiving the message $M_{A_{\text{sp}}}$, it autonomously determines which collaborators' trust should be evaluated. Agent $A_{\text{tm}}$ maintains a local trust hypergraph $\mathcal{H}_{b_i} = (B_{b_i}, E_{b_i}, W_{b_i})$ for device $b_i$, which categorizes its potential collaborators into three groups: untrusted ($e^{\text{un}}_{b_i}$), trusted-with-unstable-trend ($e^{\text{tu}}_{b_i}$), and trusted-with-stable-trend ($e^{\text{ts}}_{b_i}$), as shown in Fig.~\ref{semantic-chain}. It is worth noting that the number of groups can be dynamically extended. Each potential collaborator $b_j$ in $\mathcal{H}_{b_i}$  is associated with a set of attribute information, including its semantic trust, the timestamp of its most recent evaluation, and the number of prior evaluations. In Fig.~\ref{semantic-chain}~(b), device $b_j$ is assumed to belongs to $b_i$'s trusted-with-unstable-trend group, with attributes represented as \{``device'' : ``$b_j$'', ``latest evaluation time'' : ``2025-07-01 10:00:00'', ``cumulative evaluation count'' : ``5'', ``semantic trust'' : ``trusted-with-unstable-trend: response latency is trusted with an increasing trend, task transmission success rate is trusted with a declining trend, execution speed is trusted with a stable trend,  task computation success rate is trusted with a stable trend''\}. Agent $A_{\text{tm}}$ utilizes the latest evaluation time and the cumulative evaluation count to calculate a re-evaluation priority for each device. With this mechanism, a collaborator is given low priority if it has been evaluated recently and frequently. In contrast, devices lacking recent updates or sufficient evaluation records are assigned high immediate priority. As shown in Fig.~\ref{semantic-chain}~(b), we assume that collaborator $b_j$ is chosen for re-evaluation due to a limited number of previous evaluations and an outdated last evaluation. Agent $A_{\text{tm}}$ subsequently outputs a message $M_{A_{\text{tm}}}$ = ``The collaborator selected for trust evaluation is \{``device'' : ``$b_j$'', ``recent evaluation time'' : ``2025-07-01 10:00:00''\}".

\textit{Step 3: Agent $A_{\text{hdc}}$ collects historical collaboration data of the selected device}. Agent $A_{\text{hdc}}$ is a subscriber of the messages of agent $A_{\text{tm}}$. Upon receiving the message $M_{A_{\text{tm}}}$, agent $A_{\text{hdc}}$ formulates a new message $M_{\text{his}}$ = ``I would like to know whether you have device $b_j$'s historical collaboration data after 2025-07-01 10:00:00.'' This message is then broadcast to other devices in the system via the communication module. Upon receiving the broadcast, devices in a busy state do not respond. Devices in an idle state search their locally stored records based on the specified device and time. If relevant records are found, the devices promptly respond with the data; otherwise, they withhold any reply. We assume that devices $b_m$, $b_k$ and $b_n$ obtain their respective historical collaboration records associated with device $b_j$--namely $d_{b_m \to b_j}$, $d_{b_k \to b_j}$, and $d_{b_n \to b_j}$. They subsequently transmit these records to device $b_i$. After receiving responses, agent $A_{\text{hdc}}$ consolidates them into a structured message $M_{A_{\text{hdc}}}$ = \{``$d_{b_m \to b_j}$" : ``\{``device" : ``$b_j$", ``time" : ``2025-07-02 12:10:00", ``response latency" : ``0.5 second", ``task transmission success rate" : ``100\%", ``execution speed" : ``2 MB/s", ``task computation success rate" : ``100\%"\}", ``$d_{b_k \to b_j}$" : ``\{ ``device" : ``$b_j$", ``time" : ``2025-07-02 14:10:12", ``response latency" : ``0.4 second", ``task transmission success rate" : ``100\%", ``execution speed" : ``2.5 MB/s", ``task computation success rate" : ``100\%" \}", ``$d_{b_n \to b_j}$" : ``\{ ``device" : ``$b_j$", ``time" : ``2025-07-03 18:10:00", ``response latency" : ``0.5 second", ``task transmission success rate" : ``100\%", ``execution speed" : ``2 MB/s", ``task computation success rate" : ``100\%" \}"\}.

{\textit{Step 4: Agent $A_{\text{hte}}$ evaluates the historical collaboration-based trust of the selected device}}. Agent $A_{\text{hte}}$ is a subscriber to the messages sent by agent $A_{\text{hdc}}$. Upon receiving the message $M_{A_{\text{hdc}}}$, agent $A_{\text{hte}}$ constructs a time-ordered sequence for each of the four dimensions: response latency, task transmission success rate, execution speed, and task computation success rate. Since agent $A_{\text{hte}}$ is initialized with domain-specific knowledge in communications and computing, it can leverage this expertise to analyze the sequences to determine whether the values exhibit abnormalities and whether their temporal trends deviate from expected behavior. For instance, the response latency sequence $\{0.5, 0.4, 0.5\}$ is evaluated as trusted with a stable trend. Agent $A_{\text{hte}}$ integrates the evaluation results from the four dimensions to generate the semantic trust for device $b_j$, and produce a message $M_{A_{\text{hte}}}$ containing the timestamp of this evaluation, represented as \{``device'' : ``$b_j$'', ``evaluation time'' : ``2025-07-04 00:00:00'', ``semantic trust'' : ``trusted-with-stable-trend: response latency is trusted with a stable trend, task transmission success rate is trusted with a stable trend, execution speed is trusted with a stable trend,  task computation success rate is trusted with a stable trend''\}.



{\textit{Step 5: Agent $A_{\text{tm}}$ updates the local trust hypergraph $\mathcal{H}_{b_i}$}}. Agent $A_{\text{tm}}$ subscribes to messages published by agent $A_{\text{hte}}$. Upon receiving the message $M_{A_{\text{hte}}}$, agent $A_{\text{tm}}$ updates the local trust hypergraph in accordance with the current semantic trust of the device under evaluation. Agent $A_{\text{tm}}$ detects the current semantic trust of device $b_j$ differs from its previous semantic trust. As a result, device $b_j$ is moved from the trusted-with-unstable-trend group to the trusted-with-stable-trend group, as shown in Fig.~\ref{semantic-chain}~(e). Device $b_j$'s attributes are updated to \{``device'' : ``$b_j$'', ``latest evaluation time'' : ``2025-07-04 00:00:00'', ``cumulative evaluation count'' : ``6'', ``semantic trust'' : ``trusted-with-stable-trend: response latency is trusted with a stable trend, task transmission success rate is trusted with a stable trend, execution speed is trusted with a stable trend,  task computation success rate is trusted with a stable trend''\}.

All devices perform the above five steps during their idle periods to evaluate the semantic trust of their potential collaborators based on historical collaborations. Each device maintains a local trust hypergraph that establishes default connections among devices according to semantic trust. The trusted-with-stable-trend groups--e.g., $e^{\text{ts}}_{b_i}$ for device $b_i$, $e^{\text{ts}}_{b_m}$ for device $b_m$, and $e^{\text{ts}}_{b_k}$ for device $b_k$--are linked in a chain-like manner to form the global hypergraph $\mathcal{H}$.


\textbf{2) Resource-Based Trust Evaluation for Collaborator Selection}

When the task owner $b_i$ initiates the task $C$, it needs to select a trustworthy collaborator for task execution. To accelerate the collaborator selection process, $b_i$ does not need to re-evaluate all potential collaborators based on their historical collaborations. Instead, it directly identifies a group of trusted devices from its local trust hypergraph and then performs a task-specific resource trust evaluation on this filtered group. The most trustworthy collaborator is subsequently chosen from the evaluated candidates. The procedure is presented as follows.

{\textit{Step 6: Agent $A_{\text{tm}}$ identifies the potential collaborators.}} Upon receiving the task generation notification, $A_{\text{tm}}$ extracts all collaborators in the trusted-with-stable-trend group $e^{\text{ts}}_{b_i}$ from the local trust hypergraph $\mathcal{H}_{b_i}$. As shown in Fig.~\ref{semantic-chain}~(e), devices $b_m$ and $b_j$ are selected as the potential collaborators. Then, agent $A_{\text{tm}}$ combines the task $C$ with the selected potential collaborators to generate a message $M_{\text{task}}$: ``The task $C$ is \{$\cdots$\}. The potential collaborators are $b_m$ and $b_j$".

{\textit{Step 7: Agent $A_{\text{rdc}}$ collects the potential collaborators' resource information}}. Upon receiving the message $M_{\text{task}}$, agent $A_{\text{rdc}}$, as a subscriber to messages from agent $A_{\text{tm}}$, first identifies the task’s required resource dimensions. It then gathers the relevant resource information from the potential collaborators according to these requirements. After collecting the collaborators' resource information, agent $A_{\text{rdc}}$ organizes the resource data together with the task into a message $M_{A_{\text{rdc}}}$ = \{``task $C$" : ``$\cdots$", ``$b_m$" : ``\{``cooperative willingness" : ``yes", ``available storage" : ``20 GB", ``available CPU" : ``3 GHz"\}", ``$b_j$" : ``\{``cooperative willingness" : ``yes", ``available storage" : ``10 GB", ``available CPU" : ``1 GHz"\}" \}. This message is then sent to its subscriber.

{\textit{Step 8: Agent $A_{\text{rte}}$ evaluates the potential collaborators’ resource trustworthiness.}} 
Agent $A_{\text{rte}}$ subscribes to messages published by agent $A_{\text{rdc}}$. Upon receiving $M_{A_{\text{rdc}}}$, it initiates the resource evaluation. The evaluation of a device’s resource trustworthiness aims to determine whether its resources can meet the requirements of a specific task. In this study, as the primary focus is on presenting the proposed model, only three dimensions--cooperative willingness, available storage, and available CPU--are considered. However, in practical applications, task requirements may span multiple resource dimensions, such as computing power, storage capacity, communication bandwidth, and energy constraints.  Moreover, different tasks impose varying demands on resources, significantly increasing the complexity of the evaluation process. To achieve an accurate assessment of collaborators’ resources in dynamic tasks, this study leverages the powerful learning capabilities of agentic AI driven by the large AI model, combined with a few-shot learning strategy. As shown in Fig.~\ref{semantic-chain}~(g), historical tasks and the corresponding evaluations of their collaborators are first fed into the agent $A_{\text{rte}}$, enabling it to learn the rules and patterns of resource assessment from limited samples. Subsequently, for the new task $C$,  $A_{\text{rte}}$ evaluates the resources of potential collaborators based on the learned rules, producing reliable judgments. Through few-shot learning, $A_{\text{rte}}$ can quickly adapt to new tasks in dynamic environments, achieving efficient and generalizable assessment of resource trustworthiness. Finally, $A_{\text{rte}}$ outputs the resource evaluation results as \{``$b_m$" : ``\{cooperative willingness : yes, satisfies that collaborator willingness must be yes; available storage : 20 GB, exceeds the requirement of 500 MB; available CPU : 3 GHz, exceeds the minimum requirement of 2 GHz; It can fulfill the task $C$'s resource requirements.\}", ``$b_j$" : ``\{cooperative willingness : yes, satisfies that collaborator willingness must be yes; available storage : 10 GB, exceeds the requirement of 500 MB; available CPU : 1 GHz, below the minimum requirement of 2 GHz; It cannot satisfy the task $C$'s resource requirements.\}"\}. Therefore, device $b_m$ is selected as the final collaborator, as it demonstrates high trustworthiness in historical collaboration and provides sufficient resources for the current task.

If the resources of both $b_m$ and $b_j$ satisfy the task requirements, the task owner chooses either one. If neither device provides adequate resources, the task owner $b_i$ can leverage the global hypergraph $\mathcal{H}$ to quickly identify a remote trusted collaborator. Specifically, $b_i$ notifies devices $b_m$ and $b_j$ to initiate the collaborator selection process from their respective trusted-with-stable-trend groups. For example, the agents of device $b_m$ execute steps 6-8 to evaluate the resources of devices $b_k$ and $b_q$ within device $b_m$'s trusted-with-stable group $e^{\text{ts}}_{b_m}$. If device $b_k$ meets the resource requirements of task $C$, it is selected as the final collaborator. Then, the task $C$ is forwarded to device $b_k$ through the intermediate device $b_m$. If no device meets the task's resource requirements, the process is repeated in the global hypergraph until a trusted collaborator is found.

The proposed semantic chain-of-trust establishes an intelligent trust evaluation orchestration. It fully utilizes device idle periods for historical collaboration-based trust evaluation, while the task owner only conducts task-specific resource trust evaluation when a task arrives, significantly reducing communication and computational overhead.

\begin{figure}[!]
\centering
\includegraphics[scale=0.7]{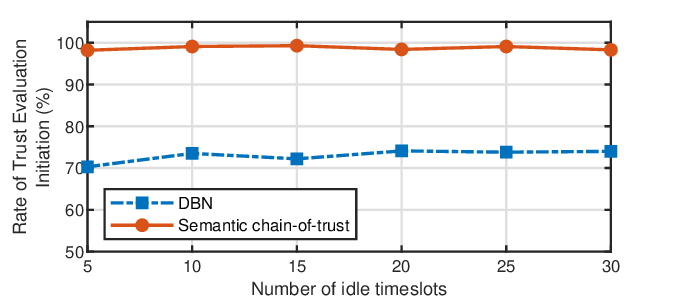}
\caption{{The proposed semantic chain-of-trust substantially improves accuracy in detecting idle states and initiating trust evaluations.}}
\label{proportion}
\end{figure}

\section{Experimental analysis}
To validate the proposed semantic chain-of-trust, we use a set of MacBook, DELL 5280, DELL EMC 5200, and DELL 3910 devices to form a collaborative system. All devices are interconnected via Wi-Fi and equipped with face recognition software. The task parameters are set by default as follows: a size of 500 MB, a minimum CPU requirement of 2 GHz, and a collaborator willingness requirement set to yes.

The validation first focuses on the accuracy of triggering trust evaluation operations during idle states. The experiment is conducted on a MacBook, where the device alternates between idle and busy periods by performing operations or remaining inactive. Durations for both types of periods are drawn from exponential distributions. The proposed semantic chain-of-trust is then compared with the Dynamic Bayesian Network (DBN)~\cite{4658625}, as shown in Fig.~\ref{proportion}. The proposed approach achieves a higher trust evaluation initiation rate during idle slots than DBN, demonstrating its superior accuracy in identifying device idle states and thus enabling more effective utilization of device resources.

\begin{figure}[!]
\centering
\includegraphics[scale=0.55]{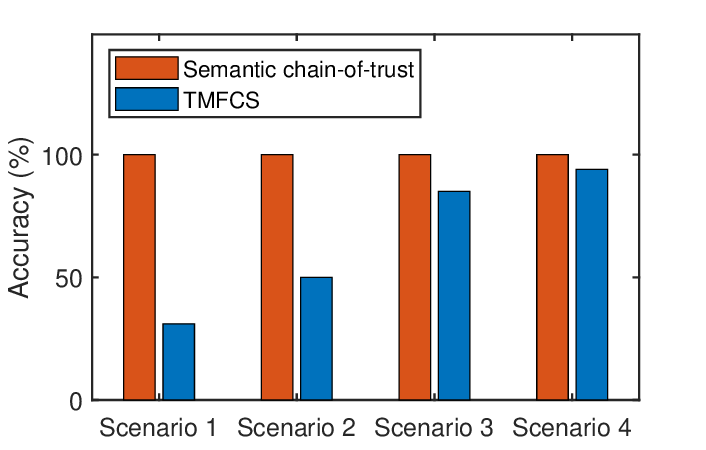}
\caption{The proposed semantic chain-of-trust achieves 100\% accuracy in identifying untrusted devices across all four scenarios.}
\label{scenario}
\end{figure}

The accuracy of the proposed semantic chain-of-trust in evaluating device trustworthiness based on historical collaborations is further assessed. Considering four dimensions of semantic trust, four scenarios are designed: in Scenario 1, 20\% of the devices have anomalous response latency; in Scenario 2, 20\% of the devices have anomalies in both response latency and task transmission; in Scenario 3, 20\% of the devices have anomalies in response latency, task transmission, and execution speed; and in Scenario 4, 20\% of the devices have anomalies in all four dimensions. The accuracy of untrusted device identification by the proposed model is then compared with that of the trust model with fitness-based clustering scheme (TMFCS)~\cite{10944812}. As shown in Fig.~\ref{scenario}, across all scenarios, the semantic chain-of-trust achieves 100\% accuracy in identifying untrustworthy devices, which demonstrates the advantage of the large AI model-driven agent in detecting anomalous behavior. In contrast, the TMFCS algorithm only attains a high identification rate in Scenario 4, where multiple dimensions are simultaneously anomalous.

\begin{figure}[!t]
      \centering
      \subfigure[]{\includegraphics[scale=0.54]{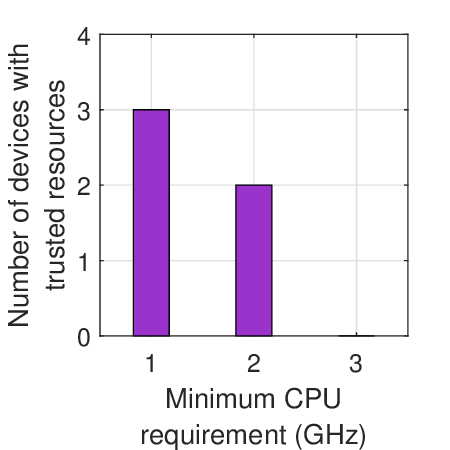}}
      \hspace{-0.01 in}\subfigure[]{\includegraphics[scale=0.54]{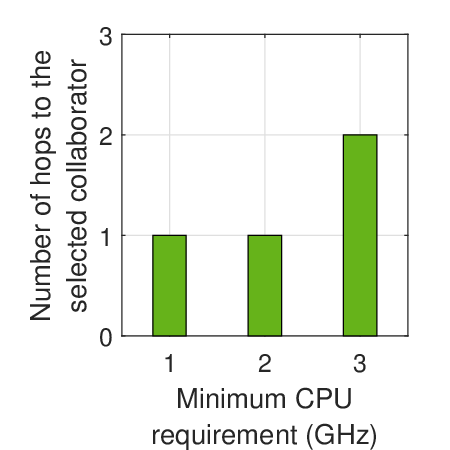}}
      \caption{Impact of task requirements on collaborator selection. (a) The number of resource-trusted collaborators within the task owner's trusted-with-stable-trend group decreases as the minimum CPU requirement increases. (b) The task owner needs more than one hop to locate a resource-trusted collaborator as the minimum CPU requirement rises.}
     \label{resource}
\end{figure}


We further examine the impact of varying task requirements on trusted collaborator selection.
A MacBook serves as the task owner, and the minimum CPU requirement is gradually increased to observe the results. In Fig.~\ref{resource}~(a), the number of devices in the task owner's trusted-with-stable-trend group that satisfy the resource requirements decreases as the minimum CPU requirement rises. When the minimum CPU requirement does not exceed 2 GHz, the task owner can still find a resource-trusted collaborator in the group, i.e., one hop, as shown in Fig.~\ref{resource}~(b). However, when the minimum CPU requirement reaches 3 GHz, no device in the group meets the resource criteria, and the task owner needs two hops to locate a trusted collaborator.

\section{Conclusion}
In this research, we have proposed the concept of semantic trust to provide a more accurate, multidimensional representation of device trustworthiness. To enable effective collaborator selection, we have further proposed a semantic chain-of-trust model and realized this model through the integration of agentic AI and hypergraphs. By deploying an agentic AI system on each device, collaborators’ semantic trust can be autonomously evaluated through a series of intelligent operations. Devices autonomously establish semantic trust-based links through hypergraphs, enabling efficient one-hop and multi-hop collaborator selection. The proposed semantic chain-of-trust establishes a foundation for autonomous, trust-aware collaboration in dynamic networks, paving the way for intelligent, scalable, and reliable multi-device systems.



\footnotesize

\end{document}